\documentclass{article}

\usepackage[numbers]{natbib}

\usepackage[final]{nips_2017}

\usepackage[utf8]{inputenc} 
\usepackage[T1]{fontenc}    
\usepackage{url}            
\usepackage{booktabs}       
\usepackage{amsfonts}       
\usepackage{nicefrac}       
\usepackage{microtype}      

\usepackage{amsmath}
\usepackage{amssymb}
\usepackage[inline]{enumitem}
\usepackage{graphicx}
\usepackage{algorithm}
\usepackage[noend]{algorithmic}
\usepackage{color}
\usepackage{subcaption}
\usepackage[super]{nth}
\usepackage{siunitx}

\title{Learning Active Learning from Data}
\author{
  Ksenia Konyushkova \\
  EPFL\\
  \texttt{ksneia.konyushkova@epfl.ch} \\
  \And
  Raphael Sznitman \\
  University of Bern \\
  \texttt{raphael.sznitman@artorg.unibe.ch} \\
  \AND
  Pascal Fua \\
  EPFL \\
  \texttt{pascal.fua@epfl.ch} \\
}
\newif\ifdraft
\drafttrue

\ifdraft
  
  \newcommand{\PF}[1]{\textcolor{red}{{\bf PF: #1}}}
  
  \newcommand{\RS}[1]{\textcolor{blue}{{\bf RS: #1}}}

\else
  \newcommand{\PF}[1]{}
  \newcommand{\RS}[1]{}
  \newcommand{\KK}[1]{}

\fi

\newcommand{\random}{{\bf Rs}}
\newcommand{\uncertainty}{{\bf Us}}
\newcommand{\kapoor}{{\bf Kapoor}}
\newcommand{\ALBE}{{\bf ALBE}}
\newcommand{\randLAL}{{\bf LAL-independent-2D}}
\newcommand{\iterLAL}{{\bf LAL-iterative-2D}}
\newcommand{\bigstartLAL}{{\bf LAL-independent-WS}}

\newcommand{\LALdatageneration}{{\textsc{DataMonteCarlo}}}
\newcommand{\funsplit}{{\textsc{Split}}}
\newcommand{\LALrandonstrategy}{{\textsc{buildLALindependent}}}
\newcommand{\LALiterativestrategy}{{\textsc{buildLALiterative}}}

\newcommand{\LALindepend}{{\textsc{LALindependent}}}
\newcommand{\LALiterative}{{\textsc{LALiterative}}}

\newcommand{\LAL}{{\textsc{LAL}}}

\DeclareMathOperator*{\argmax}{arg\,max}

\begin{document}

\maketitle

\graphicspath{{images/}}

\begin{abstract}

In this paper, we suggest a novel data-driven approach to active learning (AL). 
The key idea is to train a regressor that predicts the expected error reduction for a candidate sample in a particular learning state. 
By formulating the query selection procedure as a regression problem we are not restricted to working with existing AL heuristics; instead, we learn strategies based on experience from previous AL outcomes. 
We show that a strategy can  be learnt either from simple synthetic 2D datasets or  from a subset of domain-specific data. 
Our method yields strategies that work well on real data from a wide range of domains. 

\end{abstract}


\section{Introduction}
\label{sec:introduction}

Many modern machine  learning techniques  require large amounts  of training  data to reach   their   full   potential. 
However, annotated data is hard and expensive to obtain, notably in specialized domains where only experts whose time is scarce and precious can provide reliable labels. Active learning (AL) aims to ease the data collection process by automatically deciding  which instances an annotator should label to train an algorithm as quickly and effectively as possible. 

Over  the  years  many  AL  strategies  have  been  developed  for various classification tasks,  without any  one of  them clearly  outperforming others in all cases. 
Consequently, a number of meta-AL approaches have been proposed to automatically select the best strategy. Recent examples include bandit algorithms~\cite{Baram04,Hsu15, Chu16b} and reinforcement learning approaches~\cite{Ebert12}.
A common  limitation of these methods  is that they  cannot go beyond combining pre-existing hand-designed heuristics. 
Besides, they require reliable assessment of the classification performance which is problematic because the annotated data is scarce.
In  this paper,  we overcome these limitations thanks to two features of our approach.
First, we look at a whole continuum of AL strategies instead of combinations of pre-specified heuristics.
Second, we bypass the need to evaluate the classification quality from application-specific data because we rely on experience from previous tasks instead.

More specifically, we formulate Learning  Active Learning (\LAL{}) as a regression problem. 
Given a trained classifier and its output for a specific sample without a label, we  predict the  reduction  in  generalization error  that  can  be expected  by adding the label to that point. 
In practice, we show that we can train this regression function on synthetic data by using simple features, such as the variance of the classifier output or the predicted probability distribution over possible labels for a specific datapoint. 
Furthermore, if a sufficiently large annotated set can be provided initially, the regressor can be trained on it instead of on synthetic data.
The resulting AL strategy is then tailored to the particular problem at hand, and can be used to further extend the initial dataset.
We   show that  \LAL{}  works  well  on  real data  from  several
different  domains such  as biomedical  imaging, economics,  molecular
biology and high energy physics. 
This query selection strategy outperforms competing methods without requiring hand-crafted heuristics and at a comparatively low computational cost. 


\section{Related work }
\label{sec:related}

The extensive development of AL in the last decade has resulted in various AL strategies.
They include uncertainty sampling~\cite{Tong02,Joshi09,Settles10,Yang15}, query-by-committee~\cite{GiladBachrach05,Iglesias11},      expected model change~\cite{Settles10,Sznitman10,Vezhnevets12}, expected error or variance  minimization~\cite{Joshi12, Hoi06} and information gain~\cite{Houlsby11}.
Among  these, uncertainty  sampling  is  both simple and computationally efficient. 
This makes it one of  the  most popular strategies in real applications. 
In short, it suggests  labeling  samples that  are  the  most  uncertain,  i.e.,  closest  the classifier's decision boundary. 
The above methods work  very   well   in   cases such as the ones depicted in the top row of Fig.~\ref{fig:synthetic}, but  often fail  in the  more difficult  ones of the bottom row~\cite{Baram04}.

Among AL methods, some cater to specific    classifiers,    such    as    those relying on    Gaussian Processes~\cite{Kapoor07},   or to specific applications, such as natural language processing~\cite{Tong02,Olsson09},  sequence   labeling  tasks~\cite{Settles08b}, visual recognition~\cite{Luo04,Long15a}, semantic segmentation~\cite{Vezhnevets12}, foreground-background segmentation~\cite{Konyushkova15}, and preference learning \cite{singla2016a, Maystre17}. 
Moreover, various query  strategies aim  to maximize  different performance  metrics, as evidenced in the case of multi-class classification~\cite{Settles10}.
However, there is no one algorithm that consistently outperforms all others in all applications~\cite{Settles08b}. 

Meta-learning algorithms have been gaining in popularity in recent years~\cite{tamar2016, Santoro16}, but few AL scenarios tackle the problem of learning AL strategies.
\citet{Baram04} combine several known heuristics with the help of a bandit algorithm. 
This is made possible by the maximum entropy criterion, which estimates the classification performance without labels.
\citet{Hsu15} improve it by moving the focus from datasamples as arms to heuristics as arms in the bandit and use a new unbiased estimator of the test error.
\citet{Chu16b} go further and transfer the bandit-learnt combination of AL heuristics between different tasks.
Another approach is introduced by~\citet{Ebert12}. It involves balancing exploration and exploitation in the choice of samples with a Markov decision process.

The two main  limitations of  these  approaches  are as follows.  
First, they  are restricted to combining  already existing  techniques and second, their success depends on the ability to estimate the classification performance from scarce data.
The data-driven nature of \LAL{} helps to overcome these limitations.
Sec.~\ref{sec:experiments} shows that it outperforms several baselines including those of \citet{Hsu15} and \citet{Kapoor07}.
The method of \citet{Hsu15} is chosen as a our main baseline because it is a recent example of meta AL and is known to outperform several benchmarks.

\section{Towards data-driven active learning}
\label{sec:motivation}

In this section  we briefly introduce the active leaning framework along with uncertainty sampling (US), the most frequently-used AL heuristic. 
Then, we motivate why a data-driven approach can improve AL strategies and how it can deal with the situations where US fails.
We selected US as a representative method because it is popular and widely applicable, however the behavior that we describe is not specific to this strategy.

\subsection{Active learning (AL)}
\label{sec:standardAL}

Given a machine learning model and a pool of unlabeled data, 
the goal of AL  is to select which data should be annotated in order  to learn the  model as quickly as  possible.  
In practice,  this means that instead  of asking experts  to annotate all  the data,  we select iteratively and adaptively which  datapoints should be  annotated next.
In this paper we are interested in classifying datapoints from a target dataset $\mathcal{Z} = \{(x_1, y_1),\ldots,(x_N, y_N)\}$, where $x_i$ is a $D$-dimensional feature vector and $y_i \in \{0,1\}$ is its binary label. 
We  choose   a probabilistic classifier   $f$  that  can  be   trained  on  some $\mathcal{L}_t  \subset  \mathcal{Z}$  to  map  features  to labels, $f_t(x_i)  = \hat{y}_i$,  through  the predicted    probability   $p_t(y_i=y    \mid x_i)$.
The standard AL procedure unfolds as follows.
\begin{enumerate}
  
\item The algorithm  starts with  a small labeled  training dataset  $\mathcal{L}_t \subset \mathcal{Z}$ and large pool of annotated data $\mathcal{U}_t = \mathcal{Z} \setminus \mathcal{L}_t$ with $t=0$.
  
\item A classifier $f_t$ is trained using $\mathcal{L}_t$.
  
  
\item A query  selection procedure picks an instance $x^*  \in \mathcal{U}_t$ to be annotated  at the next  iteration.  
  
\item $x^*$  is given  a label  $y^*$ by an oracle. The labeled and unlabeled sets are updated.

\item $t$ is incremented, and steps \num{2}--\num{5} iterate until the desired accuracy is   achieved or the number of iterations has reached a predefined limit.

\end{enumerate}

\paragraph{Uncertainty sampling (US)}
\label{sec:uncertainty}

US  has been  reported to  be successful  in numerous  scenarios and settings and despite its simplicity, it often works  remarkably well~\cite{Tong02,Joshi09,Settles10,Yang15,Konyushkova15,Mosinska16}.
It focuses its selection on samples which the  current classifier  is the least  certain about. 
There are  several definitions of maximum uncertainty but one of the most widely used ones is  to select a sample $x^*$ that  maximizes the  entropy $\mathcal{H}$  over the predicted classes:
\begin{equation}
x^* = \argmax_{{x_i} \in \mathcal{U}_t}  \mathcal{H}[p_t(y_i=y \mid x_i)]
\; .
\label{eq:uncertainty}
\end{equation}

\subsection{Success, failure, and motivation}
\label{subsec:motivational-example}

\begin{figure}[h]
\includegraphics{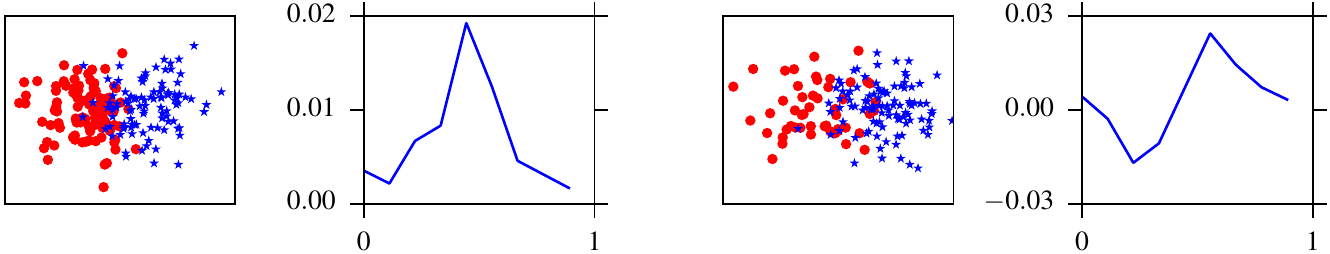}
\caption{Balanced vs unbalanced.  Left: two Gaussian  clouds of the same size. 
Right: two Gaussian  clouds with the  class $0$  twice bigger
  than class  $1$.  The test error reduction as a
  function  of predicted  probability  of class  $0$ in the respective datasets. }
   \label{fig:exampleGauss}
\end{figure}

We now motivate the need for LAL by presenting two toy  examples. 
In the first one, US is empirically observed to be the best greedy approach, but in the second it makes suboptimal decisions. 
Let us consider simple  two-dimensional datasets  $\mathcal{Z}$ and $\mathcal{Z}'$ drawn from the same distribution
with an equal number of points in each class (Fig.~\ref{fig:exampleGauss}, left).
The data in each  class comes from a Gaussian distribution with a different mean and the same variance.
We  can  initialize  the  AL procedure of Sec.~\ref{sec:standardAL} with one sample from each class and its    respective label: $\mathcal{L}_0=\{(x_1,0), (x_2,1) \} \subset \mathcal{Z}$ and $\mathcal{U}_0 = \mathcal{Z} \setminus \mathcal{L}_0$. 
Here we train a simple logistic  regression classifier  $f$  on  $\mathcal{L}_0$ and  then  test it  on $\mathcal{Z}'$. 
If~ $|\mathcal{Z}'|$ is large, the test error can be considered as a good  approximation of the generalization error:  
$\ell_0 = \sum_{(x',y') \in \mathcal{Z}'}\ell(\hat{y},y')$, where $\hat{y}=f_0(x')$.
Let  us try to label  every point $x$ from~ $\mathcal{U}_0$ one  by  one,  form a  new  labeled  set $\mathcal{L}_x  = \mathcal{L}_0   \cup   (x,y)$   and   check  what   error   a   new   classifier $f_x$  yields on  $\mathcal{Z}'$, that  is, $\ell_x  = \sum_{(x',y') \in  \mathcal{Z}'} \ell (\hat{y},y')$, where $\hat{y}=f_x(x')$.   
The  difference between  errors  obtained with classifiers  constructed on  $\mathcal{L}_0$ and  $\mathcal{L}_x$ indicates how  much the addition  of a  new datapoint  $x$ reduces  the generalization  error: $\delta_x = \ell_0-\ell_x$.
We plot $\delta_x$ for the $0 / 1$ loss function, averaged over \num{10000} experiments as a function of the predicted probability $p_0$ (Fig.~\ref{fig:exampleGauss}, left). 
By design, US would select a  datapoint with probability of class \num{0} close to \num{0.5}.
We observe that in this experiment, the datasample with $p_0$ closest to \num{0.5} is indeed the one that yields the greatest error reduction. 

In the next experiment, the class \num{0} contains twice as many datapoints as the other class, see Fig.~\ref{fig:exampleGauss}, right. 
As  before, we plot the  average error reduction as a function of $p_0$ in Fig.~\ref{fig:exampleGauss} (right). 
We observe this time that the value of  $p_0$ that  corresponds to  the largest  expected error  reduction is different from \num{0.5} and thus the choice of US becomes suboptimal. 
Also, the reduction in error is no longer symmetric for the two classes.
The more imbalanced  the two classes are, the further from the optimum the choice made by US is. 
In complex realistic scenario, there are many  other factors  such as label noise, outliers or shape of distribution that further compound the problem.

Although query  selection procedures can  take  into account  statistical properties of the  datasets and classifier, there is  no simple way to foresee the influence of all possible  factors.  
Thus, in this paper, we suggest Learning Active Learning (\LAL{}). 
It uses properties of classifiers and data to predict the potential  error reduction.   
We treat  the query  selection problem by using a regression model; this perspective enables us to construct new AL strategies in a flexible way.       
For     instance,    in    the     example    of Fig.~\ref{fig:exampleGauss} (right)  we  expect  \LAL{}  to learn a model that automatically adapts its selection to the  relative prevalence  of the two  classes without  having to explicitly state such a  rule. 
  


\section{Monte-Carlo LAL}
\label{sec:approach}

Our approach to AL is data-driven and can be formulated as a regression problem.
Given a {\it representative} dataset with ground truth, we simulate an online learning procedure using a Monte-Carlo approach.
We propose two versions of AL strategies.
When building the first one, \LALindepend{}, we incorporate unused labels individually and at random to retrain the classifier.
Our goal is to correlate the change in test performance with the properties of the classifier and of newly added datapoint.
To build the \LALiterative{} strategy, we further extend our method by a sequential procedure to account for selection bias caused by AL.
We formalize our LAL procedure in the remainder of the section. 

\subsection{Independent LAL}
\label{sec:monteCarlo}

Let the  {\it representative} dataset be  split into a  training $\mathcal{D}$  and  a testing  set  $\mathcal{D}'$.
Let  $f$ be  a  classifier with a given training procedure. 
We start collecting data for the regressor by splitting   $\mathcal{D}$  into  a   labeled  set $\mathcal{L}_{\tau}$  of size $\tau$ and  an unlabeled set~ $\mathcal{U}_{\tau}$  containing the remaining points (Alg.~\ref{alg:lal-construct-data} \LALdatageneration{}).
We  then  train a  classifier  $f$  on $\mathcal{L}_{\tau}$, resulting  in  a function  $f_{\tau}$  that we  use  to  predict class  labels  for elements  $x'$  from  the  test set  $\mathcal{D}'$  and  estimate  the  test classification loss $\ell_{\tau}$.
We  characterize  the  classifier   state  by $K$ parameters $\phi_\tau = \{ \phi^1_\tau,\ldots,\phi^K_\tau \}$, which are specific to the particular  classifier type and are sensitive to the change in the training dataset while being relatively invariant to the stochasticity of the optimization procedure.  
For example,  they  can  be  the  parameters  of  the  kernel  function  if  $f$  is kernel-based, the average  depths of  the  trees  if $f$  is  a  random forest,  or prediction variability if $f$ is an ensemble classifier.
The above steps are summarized in lines \num{3}--\num{5} of Alg.~\ref{alg:lal-construct-data}.

Next, we randomly select a  new  datapoint $x$ from  $\mathcal{U}_\tau$ which is characterized by $R$ parameters $\psi_x = \{ \psi^1_x,\ldots,\psi^R_x \}$.
For example, they can include  the predicted probability to belong to class $y$,  the distance to the closest point in  the dataset or the distance to  the closest labeled point.
We  form  a new  labeled  set $\mathcal{L}_x   =  \mathcal{L}_{\tau}   \cup  \{x\}$   and  retrain $f$ (lines \num{7}--\num{13} of Alg.~\ref{alg:lal-construct-data}).
The  new classifier  $f_x$ results in the test-set loss
$\ell_x$.  
Finally,  we record  the difference  between previous  and  new loss  $\delta_x  =
\ell_{\tau}  -  \ell_x$ which is associated  to   the learning state in which it was received.
The learning state is characterized by a vector
$ \xi_\tau^x = \begin{bmatrix} \phi^1_\tau & \cdots & \phi^K_\tau & \psi^1_x & \cdots & \psi^R_x \end{bmatrix} \in \mathbb{R}^{K+R}$,
whose elements depend both on the state  of the current classifier $f_\tau$ and on the datapoint $x$.
\begin{algorithm}[h]
   \caption{\LALdatageneration{}}
   \label{alg:lal-construct-data}
\begin{algorithmic}[1]

   \STATE {\bfseries Input:} training and test datasets $\mathcal{D}$, $\mathcal{D}'$, classification procedure $f$, partitioning function \funsplit{}, size $\tau$
   \STATE {\bfseries Initialize:} $\mathcal{L}_\tau$, $\mathcal{U}_\tau \gets$ \funsplit{}($\mathcal{D}, \tau$)
   \STATE train a classifier $f_\tau$
   \STATE estimate the test set loss $\ell_\tau$
   \STATE compute the classification state parameters $\phi \gets \{ \phi^1_\tau,\ldots,\phi^K_\tau \}$
   \FOR{$m=1$ {\bfseries to} $M$}
      \STATE select $x \in \mathcal{U}_\tau$ 
      \STATE form a new labeled dataset $\mathcal{L}_x \gets \mathcal{L}_\tau \cup \{ x \}$
      \STATE compute the datapoint parameters $\psi \gets \{ \psi^1_x,\ldots,\psi^R_x \}$
      \STATE train a classifier $f_x$
      \STATE estimate the new test loss $\ell_x$
      \STATE compute the loss reduction $\delta_x \gets \ell_\tau-\ell_x$
      \STATE $\xi_m \gets \begin{bmatrix} \phi^1_\tau & \cdots & \phi^K_\tau & \psi^1_x & \cdots & \psi^R_x \end{bmatrix}$, $\delta_m \gets \delta_x$
   \ENDFOR 
   \STATE $\Xi \gets \{ \xi_m\}$  , $\Delta \gets \{ \delta_m \}$
   \STATE {\bfseries Return:} matrix of learning states $\Xi \in \mathbb{R}^{M \times (K+R)}$, 
   vector of reductions in error $\Delta \in \mathbb{R}^M$
\end{algorithmic}
\end{algorithm}
To build an AL strategy \LALindepend{} we   repeat   the   \LALdatageneration{} procedure   for   $Q$   different   initializations $\mathcal{L}_\tau^1,  \mathcal{L}_\tau^2,\ldots,\mathcal{L}_\tau^Q$ and  $T$ various labeled subset  sizes  $\tau  =  2,\ldots,T+2$ (Alg.~\ref{alg:lal-monte-carlo} lines \num{4} and \num{5}).
For  each initialization $q$ and iteration $\tau$, we sample $M$ different datapoints $x$ each of which yields classifier/datapoint state pairs with an associated reduction in error (Alg.~\ref{alg:lal-construct-data}, line \num{13}).
This   results in a matrix $\Xi \in \mathbb{R}^{(QMT)\times(K+R)}$ of observations $\xi$ and a vector $\Delta \in \mathbb{R}^{QMT}$ of labels $\delta$ (Alg.~\ref{alg:lal-monte-carlo}, line \num{9}).

Our insight is that observations $\xi$ should lie on a smooth manifold and that similar states  of  the classifier  result in similar  behaviors when annotating similar samples.
From this, a regression function can predict  the potential error reduction of annotating a specific sample in a given classifier state. 
Line \num{10} of \LALrandonstrategy{} algorithm looks for a mapping $g:  \xi \rightarrow \delta$, which is not specific to the dataset $\mathcal{D}$, and thus can be used to detect  samples that promise the greatest increase  in classifier performance in other target domains $\mathcal{Z}$.
The resulting \LALindepend{} strategy greedily selects a datapoint with the highest potential in error reduction at iteration $t$ by taking the maximum of the value predicted by the regressor $g$:
\begin{equation}
x^* =\argmax_{x \in \mathcal{U}_t} g(\phi_t, \psi_x).
\label{eq:lal-monte-carlo}
\end{equation}

\subsection{Iterative LAL}
\label{sec:approach-iterative}


For any AL strategy  at  iteration $t>0$, the labeled set  $\mathcal{L}_t$  consists of samples selected at  previous iterations, which is clearly {\it not} random.
However, in Sec.~\ref{sec:monteCarlo} the dataset $\mathcal{D}$ is split into $\mathcal{L}_\tau$ and $\mathcal{U}_\tau$ randomly no  matter how many labeled samples $\tau$ are available.

To account for  this, we modify the approach  of Section~\ref{sec:monteCarlo} in Alg.~\ref{alg:lal-iterative} \LALiterativestrategy{}. 
Instead of partitioning the dataset $\mathcal{D}$ into $\mathcal{L}_\tau$ and $\mathcal{U}_\tau$ randomly, we suggest simulating the AL procedure which selects datapoints according to the strategy learnt on the previously collected data (Alg.~\ref{alg:lal-iterative}, line \num{10}).
It first learns a strategy $\mathcal{A}(g_2)$ based on a regression function $g_2$ which selects the most promising \nth{3} datapoint when \num{2} random points are available.
In the next iteration, it learns a strategy $\mathcal{A}(g_3)$ that selects \nth{4} datapoint given \num{2} random points and \num{1} selected by $\mathcal{A}(g_2)$ etc.
In this  way, samples at each iteration depend on the samples at the previous iteration and the  sampling bias of AL is represented in the data $\Xi, \Delta$ from which the final strategy \LALiterative{} is learnt. 

The resulting strategies \LALindepend{} and \LALiterative{} are both reasonably fast during the online steps of AL.
The offline part, generating a datasets to learn a regression function, can induce a significant computational cost depending on the parameters of the algorithm.
For this reason, \LALindepend{} is preferred to \LALiterative{} when an application-specific strategy is needed.

\begin{minipage}[t]{.45\linewidth}
\begin{algorithm}[H]
   \caption{\LALrandonstrategy{}}
   \label{alg:lal-monte-carlo}
\begin{algorithmic}[1]
   \STATE {\bfseries Input:}  iteration range $\{\tau_{\min},\ldots, \tau_{\max}\}$, classification procedure $f$
   \STATE \funsplit{} $\gets$ random partitioning function
   \STATE {\bfseries Initialize:} generate train set $\mathcal{D}$ and test dataset $\mathcal{D}'$
   \FOR{$\tau$ {\bfseries in} $\{\tau_{\min}, \ldots \tau_{\max}\}$}
      \FOR{$q=1$ {\bfseries to} $Q$}
	      \STATE $\Xi_{\tau q}, \Delta_{\tau q} \gets$ \LALdatageneration{} ($\mathcal{D}, \mathcal{D}', f,$ \funsplit{}, $\tau$)
	  \ENDFOR
   \ENDFOR   
   \STATE $\Xi, \Delta \gets \{ \Xi_{\tau q} \}, \{ \Delta_{\tau q} \}$
   \STATE train a regressor $g: \xi \rightarrow \delta $ on data $\Xi, \Delta$
   \STATE construct \LALindepend{} $\mathcal{A}(g)$: \\
   $x^*=\argmax_{{x} \in \mathcal{U}_t} g[\xi_{t,x})]$
\STATE {\bfseries Return:} \LALindepend{}
\vspace{7.65mm}
\end{algorithmic}
\end{algorithm}
\end{minipage}
\begin{minipage}[t]{.45\linewidth}
\begin{algorithm}[H]
   \caption{\LALiterativestrategy{}}
   \label{alg:lal-iterative}
\begin{algorithmic}[1]
   \STATE {\bfseries Input:} iteration range $\{ \tau_{\min}, \ldots \tau_{\max} \}$, classification procedure $f$
   \STATE \funsplit{} $\gets$ random partitioning function
   \STATE {\bfseries Initialize:} generate train set $\mathcal{D}$ and test dataset $\mathcal{D}'$
   \FOR{$\tau$ {\bfseries in} $\{ \tau_{\min},\ldots, \tau_{\max} \}$}
      \FOR{$q=1$ {\bfseries to} $Q$}
	      \STATE $\Xi_{\tau q}, \Delta_{\tau q} \gets$ \LALdatageneration{} ($\mathcal{D}, \mathcal{D}', f,$ \funsplit{}, $\tau$)
	  \ENDFOR
	  \STATE $\Xi_\tau, \Delta_\tau \gets \{\Xi_{\tau q}, \Delta_{\tau q} \}$
	  \STATE train regressor $g_\tau: \xi \rightarrow \delta$ on $\Xi_\tau, \Delta_\tau$
	  \STATE \funsplit{} $\gets$ $\mathcal{A}(g_\tau)$
   \ENDFOR 
   \STATE $\Xi, \Delta \gets \{ \Xi_\tau, \Delta_\tau \}$  
   \STATE train a regressor $g: \xi \rightarrow \delta$ on $\Xi, \Delta$
   \STATE construct \LALiterative{} $\mathcal{A}(g)$: \\
   \STATE {\bfseries Return:} \LALiterative 
\end{algorithmic}
\end{algorithm}
\end{minipage}


\section{Experiments}
\label{sec:experiments}

\paragraph{Implementation details}

\label{sec:lalimplementation}

We test AL strategies in two possible settings: 
\begin{enumerate*}[label={\alph*)}]
\item {\it cold start}, where we  start with one sample from each of two classes and
\item {\it warm start}, where a larger dataset of size $N_0 \ll N$ is available to train the initial classifier.
\end{enumerate*}
The {\it warm start scenario} is largely overloooked in the litterature, but we believe it has a significant practical interest.
Learning a classifier for a real-life application with AL rarely starts from scratch, but a small initial annotated set is provided to understand if a learning based approach is applicable at all.
While a small set is good  to provide an initial insight, a real working prototype still requires much more training.
In this situation, we can benefit from the available training data to learn a specialized AL strategy for an application.
In {\it cold start} we take the representative dataset to be a \num{2}D synthetic dataset where class-conditional data distributions are Gaussian. 

In most of the experiments, we use Random  Forest (RF)  classifiers for $f$ and a RF regressor for $g$.
The state of  the learning  process consists of the following features:
\begin{enumerate*}[label={\alph*)}]
\item predicted {\it probability} $p(y=0 | \mathcal{L}_t, x)$;
\item {\it proportion} of class \num{0} in $\mathcal{L}_t$;
\item {\it out-of-bag} cross-validated accuracy of $f_t$;
\item variance of {\it feature importances} of $f_t$;
\item {\it forest variance} computed as variance of trees' predictions on $\mathcal{U}_t$;
\item average {\it tree depth} of the forest;
\item {\it size} of $\mathcal{L}_t$. 
\end{enumerate*}
For additional implementational details, including examples of the synthetic datasets, parameters of the data generation algorithm and features in the case of GP classification, we refer to the supplementary materials.

\paragraph{Baselines and protocol}
We compare the three versions of our approach: 
\begin{enumerate*}[label={\alph*)}]
  \item \randLAL{}, \LALindepend{} strategy trained on a synthetic dataset of {\it cold start};
  \item \iterLAL{}, \LALiterative{} strategy trained on a synthetic dataset of {\it cold start};
  \item  \bigstartLAL{}, \LALindepend{} strategy trained on {\it warm start} representative data;
\end{enumerate*}
against the following baselines:
\begin{enumerate*}[label={\alph*)}]
\item \random{}, random sampling;
\item \uncertainty{}, uncertainty sampling;
\item \kapoor{}~\cite{Kapoor07}, an algorithm that balances exploration and exploitation by incorporating mean and variance estimation of the GP classifier;
\item \ALBE{}~\cite{Hsu15}, a recent example of meta-AL that adaptively uses a combination of strategies, including ~\cite{Huang10}, \uncertainty{} and \random{}.
\end{enumerate*}

In all AL experiments we select samples from a training  set and report the  classification   performance  on  an independent  test  set.   
We repeat each experiment \num{50}--\num{100} times with random permutations of training and testing splits and different initializations.
Then we report the average  test performance  as  a  function  of the number  of  labeled samples.
The performance  metrics are task-specific and include classification  accuracy,  IOU~\cite{Everingham10},  dice score~\cite{Gordillo13}, AMS score~\cite{adam15}, as well as area under the ROC curve.

\subsection{Synthetic data}

\begin{figure}[h]
\includegraphics[width=\linewidth]{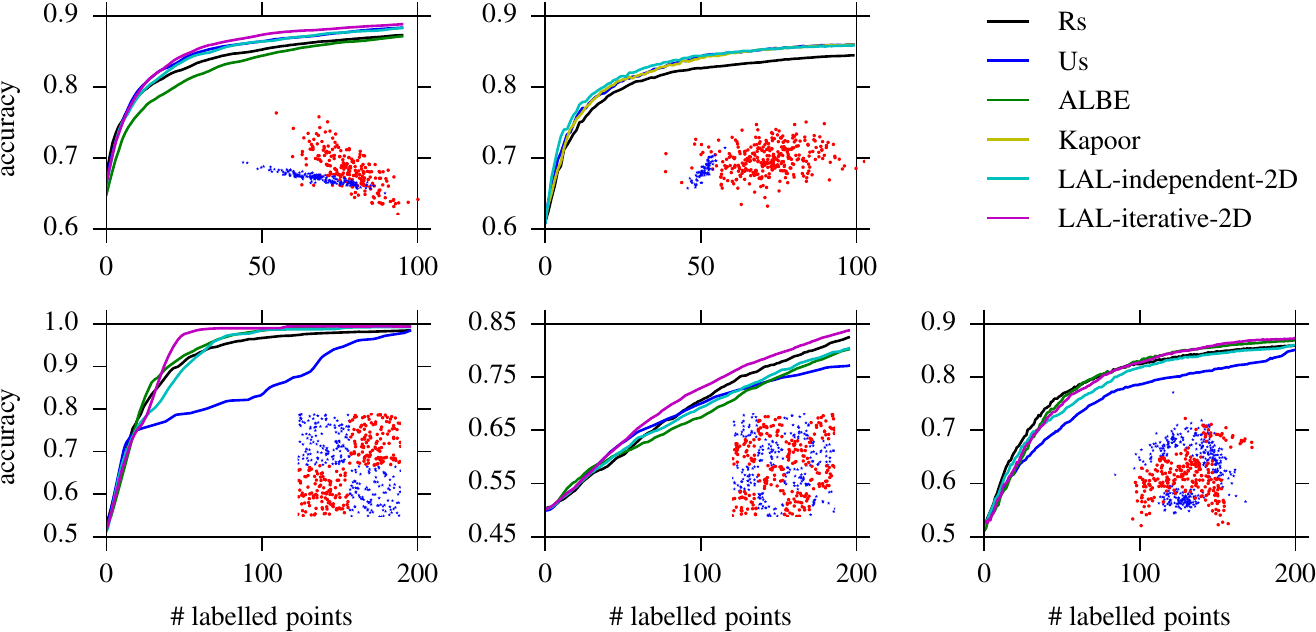}
\caption{Experiments on the synthetic data. Top row: RF and GP on 2 Gaussian clouds. Bottom row from left to right: experiments on {\it Checkerboard} \num{2 x 2}, {\it Checkerboard} \num{4 x 4}, and {\it Banana} datasets.}
\label{fig:synthetic}
\end{figure}

\paragraph{Two-Gaussian-clouds experiments}

In this dataset we test our approach with two classifiers: RF and Gaussian Process classifier (GPC).
Due to the the computational cost of GPC, it is only tested in this experiment.
We generate \num{1000} new unseen synthetic datasets as shown in the top row of Fig.~\ref{fig:synthetic}.
In both cases the proposed \LAL{} strategies selects datapoints that help to construct better classifiers faster than \random{}, \uncertainty{}, \kapoor{} and \ALBE{}.  

\paragraph{XOR-like experiments}
XOR-like datasets are known to be challenging for many  machine learning methods and AL is not an exception.
It was reported in~\citet{Baram04} that various AL algorithms struggle with tasks such as those depicted in the bottom row of Fig.~\ref{fig:synthetic}, namely {\it Checkerboard \num{2 x 2}}, {\it Checkerboard \num{4 x 4}}, and the {\it Banana} dataset from~\citet{Ratsch01}.
As previously observed, \uncertainty{}  loses to  \random{} in these cases.
\ALBE{} does not suffer from such adversarial conditions as much as \uncertainty{}, but \iterLAL{} outperforms it on {\it Checkerboard \num{2 x 2}} and {\it Checkerboard \num{2 x 2}} and matches its performance on the {\it Banana} dataset.

\subsection{Real data}

We now turn to real data from domains where annotating is hard because it requires special training to do so correctly:
\begin{enumerate*}[label={\alph*)}]
\item {\it  Striatum}, \num{3}D Electron Microscopy  stack of rat neural  tissue, the  task  is  to   detect  and  segment  mitochondria~\cite{Lucchi12, Konyushkova15};
\item  {\it  MRI}, brain  scans obtained   from the  BRATS  competition~\cite{Menze14}, the task  is to  segment brain  tumor in  T1, T2,  FLAIR, and  post-Gadolinium T1 MR images;
\item {\it Credit card}~\cite{Dal15}, a dataset of credit card transactions made in \num{2013} by  European cardholders, the task is to detect fraudulent transactions;
\item {\it Splice}, a molecular biology dataset with the task of detecting splice junctions in DNA sequences~\cite{Lorena02};
\item {\it Higgs}, a high energy physics dataset that contains measurements simulating the ATLAS experiment~\cite{adam15}, the task is to detect the Higgs boson in the noise signal.
\end{enumerate*}
Additional details about the above datasets including sizes, dimensionalities and preprocessing techniques can be found in the supplementary materials.

\begin{figure}[h]
\includegraphics[width=\linewidth]{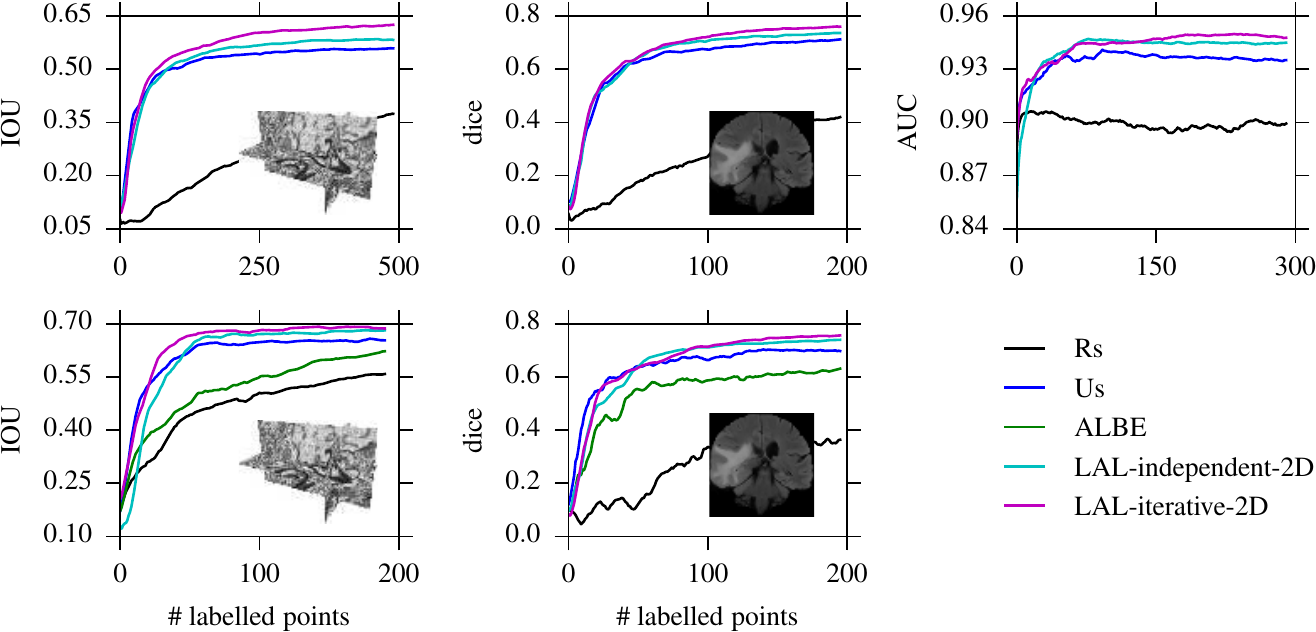}
\caption{Experiments on real data. Top row: IOU for {\it Striatum}, dice score for {\it MRI} and {AUC} for {\it Credit card} as a function of a number of labeled points. Bottom row: Comparison with \ALBE{} on the {\it Striatum mini} and {\it MRI mini} datasets.}
\label{fig:realcold}
\end{figure}

\paragraph{Cold Start AL}
Top row of Fig.~\ref{fig:realcold} depicts the  results of applying \random{}, \uncertainty{}, \randLAL{}, and \iterLAL{} on the {\it Striatum, MRI}, and {\it Credit card} datasets.  
Both \LAL{} strategies outperform \uncertainty{}, with \iterLAL{} being the best of the two. 
Considering that the \LAL{}  regressor was learned using a simple synthetic \num{2}D dataset, it  is remarkable that  it work effectively on such  complex and high-dimensional tasks.   
Due to the high computational cost of \ALBE{},  we downsample {\it Striatum} and {\it MRI} datasets to \num{2000} datapoints (referred to as {\it Striatum mini} and {\it MRI mini}).
Downsampling was not possible for the {\it Credit card} dataset due to the sparsity of positive labels (\num{0.17}\%). 
We see in the bottom row of Fig.~\ref{fig:realcold} that \ALBE{} performs even worse than \uncertainty{}. 
We ascribe this to the lack of labeled data, which \ALBE{} needs to estimate classification accuracy (see Sec.~\ref{sec:related}).

\paragraph{Warm Start AL}

\begin{figure}[h]
\includegraphics[width=\linewidth]{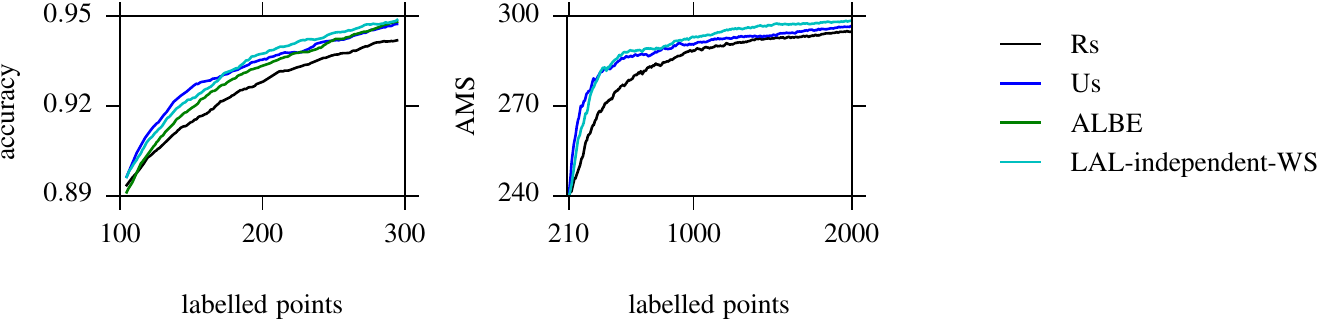}
\caption{Experiments on the real datasets with warm start. Accuracy for {\it Splice} on the left, AMS score for {\it Higgs} on the right.}
\label{fig:realwarm}
\end{figure}

In Fig.~\ref{fig:realwarm} we  compare  \bigstartLAL{}  on the  {\it Splice} and {\it Higgs} datasets by initializing \LALrandonstrategy{} with \num{100}  and \num{200}  datapoints from the corresponding tasks.
We tested \ALBE{} on the {\it Splice} dataset, however in the {\it Higgs} dataset the number of iterations in the experiment is too big for it.
\bigstartLAL{} outperforms other methods with \ALBE{} delivering competitive performance---yet, at a high computational cost---only at the end of AL.

\subsection{Analysis of LAL strategies and time comparison}
\label{sec:analysis}

\begin{figure}[h]

\begin{center}
    \includegraphics{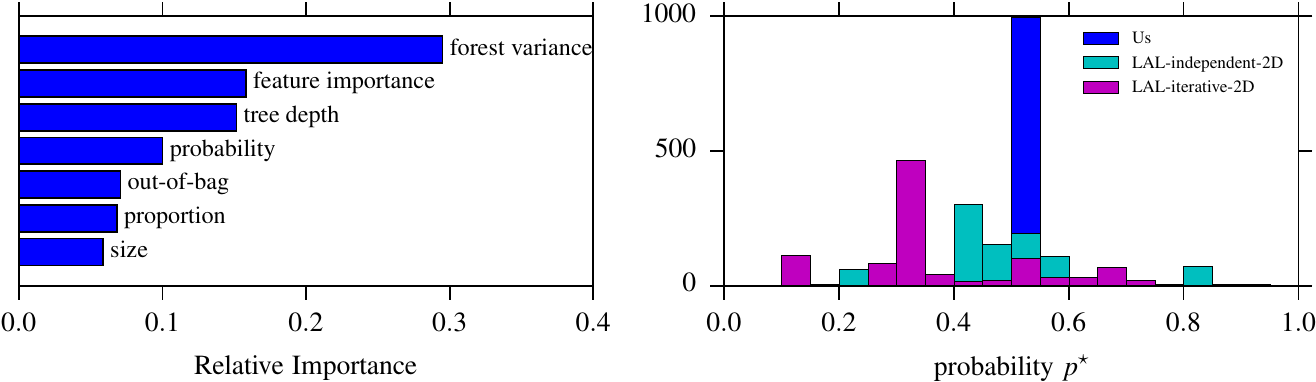}
   \caption{Left: feature importances of the RF regressor representing \LALiterative{} strategy. Right: histograms of the selected probability for different AL strategies.}
   \label{fig:featimport}
  \end{center}
\end{figure}

To better understand LAL strategies, we show in Fig.~\ref{fig:featimport} (left) the relative importance of the features of the regressor $g$ for \LALiterative{}.
As expected, both classifier state parameters and datapoint parameters influence the AL selection.
In order to understand what kind of selection \LALindepend{} and \LALiterative{} do, we record the predicted probability of the chosen datapoint $p(y^*=0 | \mathcal{D}_t, x^*)$ in \num{10} {\it cold start} experiments with the same initialization on the {\it MRI} dataset.
Fig.~\ref{fig:featimport}(right) shows the histograms of these probabilities for \uncertainty{}, \randLAL{} and \iterLAL{}.
LAL strategies have high variance and modes different from \num{0.5}.
Not only does the selection by LAL strategies differ significantly from standard US, but also the independent and iterative approaches differ from each other.
\vspace{-5mm}

\paragraph{Computational costs} 
While collecting synthetic data can be slow, it must only be done {\it once, offline,} for all applications. 
Collecting data offline for {\it warm start}, that is application specific, took us approximately \num{2.7}h and \num{1.9}h for {\it Higgs} and {\it Splice} datasets respectively. 
By contrast, the online user-interaction part is fast: it simply consists of learning $f_t$, extracting learning state parameters and evaluating the regressor $g$. 
The LAL run time depends on the parameters of the random forest regressor which are estimated via cross-validation (discussed in the supplementary materials). 
Run times of a python-based implementation with \num{1} core are given in Tab.~\ref{tab:time} for a typical parameter set ($\pm$ \num{20}\% depending on exact parameter values). 
Real-time performance can be attained by parallelising and optimising the code, even in applications with large amounts of high-dimensional data. 

\vspace{-2mm}
\begin{table}[h]
  \caption{Time in seconds for one iteration of AL for various strategies and tasks.}
  \label{tab:time}
  \label{sample-table}
  \centering
  \begin{tabular}{lrrrrr}
    \toprule
    {\bf Dataset}	& {\bf Dimensions} & {\bf \# samples} & \uncertainty{} & \ALBE{}	& \bf{\LAL{}} \\
    \midrule
    {\it Checkerboard} & \num{2}  & \num{1000} & \num{0.11} & \num{13.12} & \num{0.54} \\
    {\it MRI mini} & \num{188}  & \num{2000} & \num{0.11} & \num{64.52} & \num{0.55} \\
    {\it MRI} & \num{188}  & \num{22934} & \num{0.12} & --- & \num{0.88} \\
    {\it Striatum mini} & \num{272}  & \num{2000} & \num{0.11} & \num{75.64} & \num{0.59} \\
    {\it Striatum} & \num{272}  & \num{276130} & \num{2.05} & --- & \num{19.50} \\
    {\it Credit} & \num{30} & \num{142404} & \num{0.43} & --- & \num{4.73} \\
    \bottomrule
  \end{tabular}
\end{table}
\vspace{-5mm}


\section{Conclusion}

In this paper we introduced a new approach to AL that is driven by data: Learning Active Learning. 
We found out that Learning Active Learning from simple 2D data generalizes remarkably well to challenging new domains.
Learning from a subset of application-specific data further extends the applicability of our approach.
Finally, LAL demonstrated robustness to the choice of type of classifier and features.



\section*{Acknowledgements}

This project has received funding from the European Union’s Horizon 2020 Research and Innovation Programme under Grant Agreement No. 720270 (HBP SGA1). 
We would like to thank Carlos Becker and Helge Rhodin for their comments on the text, and Lucas Maystre for his discussions and attention to details.

\bibliographystyle{plainnat}  
\bibliography{string,learning,vision,biomed}

\end{document}